\documentclass[letterpaper]{article} 
\usepackage{aaai2026}  
\usepackage{times}  
\usepackage{helvet}  
\usepackage{courier}  
\usepackage[hyphens]{url}  
\usepackage{graphicx} 
\usepackage{natbib}  
\usepackage{caption} 
\usepackage{algorithm}
\usepackage{algorithmic}

\usepackage{caption} 
\usepackage{bm}
\usepackage{xcolor}
\usepackage{subcaption}
\usepackage{multicol}
\usepackage{multirow}
\usepackage{tabularray}
\usepackage{amsmath}
\usepackage{amssymb}
\usepackage{siunitx}
\usepackage{colortbl}
\usepackage{lipsum}
\usepackage[capitalize]{cleveref}
\usepackage{afterpage}
\usepackage{placeins}
\definecolor{clblue}{RGB}{222, 234, 246}
\definecolor{tabred}{HTML}{d62728}
\definecolor{tabblue}{HTML}{1f77b4}
\definecolor{ssim}{rgb}{0.94,0.94,1}
\definecolor{lyellow}{rgb}{1,1,0.92}
\usepackage{tabularx}
\usepackage{booktabs}

\definecolor{clblue}{RGB}{222, 234, 246}
\definecolor{purple}{HTML}{F2E6F2}

\usepackage{float}
\usepackage{newfloat}
\usepackage{listings}
\DeclareCaptionStyle{ruled}{labelfont=normalfont,labelsep=colon,strut=off} 
\floatstyle{ruled}
\newfloat{listing}{tb}{lst}{}
\floatname{listing}{Listing}
\title{Towards Optimal Aggregation of Varying Range Dependencies in Haze Removal}
\author{
    Xiaozhe Zhang, Fengying Xie, Haidong Ding, Linpeng Pan, Zhenwei Shi
}
\affiliations{
    Beihang University
}

\usepackage{bibentry}

\begin{document}

\maketitle

\begin{abstract}
Haze removal aims to restore a clear image from a hazy input. Existing methods achieve notable success by specializing in either short-range dependencies to preserve local details or long-range dependencies to capture global context. Given the complementary strengths of both, a natural progression is to explicitly integrate them within a unified framework and enable their reasonable aggregation. However, this integration remains underexplored. In this paper, we propose \textbf{\textit{DehazeMatic}}, which simultaneously and explicitly captures both short- and long-range dependencies through a dual-stream design. To optimize the contribution of dependencies at varying ranges, we conduct extensive experiments to identify key influencing factors and find that an effective aggregation mechanism should be guided by the joint consideration of haze density and semantic information. Building on these insights, we introduce the CLIP-enhanced Dual-path Aggregator, which not only enables the generation of fine-grained haze density maps for the first time, but also produces semantic maps within a shared backbone, ultimately leveraging both to instruct the aggregation process. Extensive experiments demonstrate that DehazeMatic outperforms state-of-the-art methods across multiple benchmarks.
\end{abstract}
\section{Introduction}
\label{sec:intro}
Image dehazing aims to recover a clear image from a hazy input, serving as a crucial pre-processing step for high-level vision tasks in hazy conditions, such as object detection~\cite{li2023detection} and semantic segmentation~\cite{ren2024triplane}.

Current data-driven methods can be categorized into two classes based on the receptive field size of the operators used for feature extraction, namely (\textit{\textbf{i}}) using convolution~\cite{bai2022self, cai2016dehazenet, dong2020multi, li2017aod, ren2018gated, ren2020single, zhang2018densely} or window-based self-attention~\cite{kulkarni2022unified, kulkarni2023aerial, song2023vision, wang2023uscformer} and (\textit{\textbf{ii}}) using linear self-attention~\cite{qiu2023mb} or State Space Model (SSM)~\cite{shen2023mutual,zhou2024rsdehamba,zhang2024lmhaze}. The former captures short-range dependencies, which provide fine-grained local perception but lack effective global modeling capabilities~\cite{kim2023dead,veit2016residual, de2020batch}.  In contrast, the latter captures long-range dependencies with rich contextual information while ensuring efficient computation, yet fails to preserve local 2D inductive biases~\cite{huang2024localmamba, dosovitskiy2020image}.

Both types of methods have achieved excellent performance, and their respective strengths can compensate for each other's limitations. This naturally motivates the design of a dual-stream network that explicitly integrates short- and long-range dependencies, thereby harnessing the complementary advantages of both approaches and potentially achieving superior performance. However, reasonably aggregating short- and long-range dependencies from these two paths remains a non-trivial task, as tokens with different characteristics in the image require varying ranges of dependencies. Simple operations such as addition, concatenation, or gating mechanisms~\cite{ren2018gated,zhang2020gated,hou2025bidomain} often lead to suboptimal performance. More importantly, \textit{there is still no clear guidance on how to determine the relative importance between short-range and long-range dependencies for each token}.

In this paper, we are the first to identify, through both quantitative and qualitative experiments, the key factors that determine the relative importance of short- and long-range dependencies required by each token: \textit{haze density} and \textit{semantic information}. Based on these insights, we propose a dual-stream framework for dehazing, termed \textbf{DehazeMatic}, which simultaneously captures both short- and long-range dependencies. Since feature extraction operators are not the primary focus of this work, their selection is relatively flexible. We adopt window-based self-attention~\cite{liu2021swin} and the Selective Scan Space State Sequential Model (S6) block~\cite{gu2023mamba} as representative implementations.

\begin{figure*}[ht]
    \centering
    \includegraphics[width=140mm]{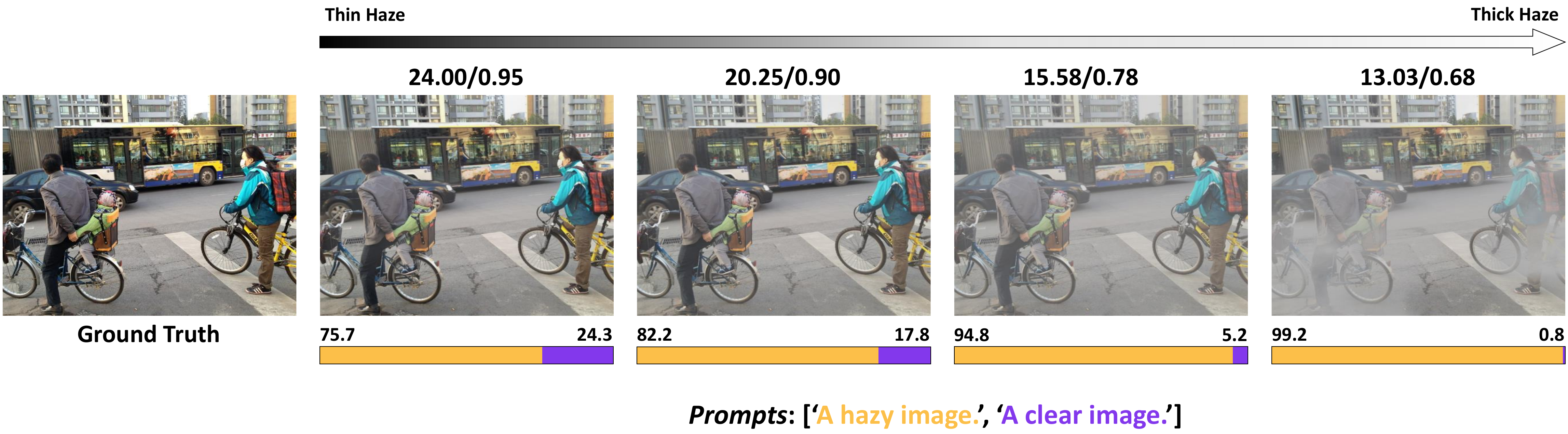}
    \vspace{-10pt}
    \caption{Illustration of CLIP~\cite{radford2021learning}'s potential to perceive haze and its density. We use the ViT-B/32 OpenCLIP~\cite{ilharco_gabriel_2021_5143773} model, pre-trained on the LAION dataset, for our experiments. The values above the images indicate the PSNR and SSIM metrics, which reflect haze density from a quantitative perspective. The values below the images represent the similarity scores between the images and the paired prompts from CLIP. Notably, as haze density increases, the similarity score with the haze-describing prompt also increases.}
    \label{fig:clip_can_show_haze_density}
    \vspace{-10pt}
\end{figure*}

Next, based on the identified key factors, we aim to generate a haze density map and a semantic information map to guide the aggregation of dependencies across varying ranges. Moreover, to improve model efficiency, we seek to generate both maps using a shared backbone. However, this objective presents significant challenges: not only is there currently no method capable of accurately estimating haze density maps, but extracting two types of information with substantial modality differences from a single backbone remains inherently difficult. Inspired by recent advances in CLIP~\cite{radford2021learning}, which is pretrained on web-scale datasets and encapsulates rich semantic priors that enable strong performance in zero-shot semantic segmentation~\cite{zhou2023zegclip,zhang2024exploring}, we further observe that CLIP has the potential to perceive haze and its density, as demonstrated in Figure~\ref{fig:clip_can_show_haze_density}. Building on this observation, we propose \textit{C}LIP-\textit{e}nhanced \textit{D}ual-path \textit{A}ggregator (CedA). Specifically, we treat the patch-wise image embeddings from the CLIP image encoder as the semantic map. In parallel, we compute affinity scores between these image embeddings and the text embeddings of paired prompts describing clear and hazy conditions—obtained from the CLIP text encoder—and use these scores as the estimated haze density map. To enable accurate estimation while avoiding labor-intensive prompt engineering, we replace manually designed prompts with learnable ones. Training proceeds in a progressive two-stage manner: we first employ cross-entropy loss, followed by regression-based training on a self-constructed triplet dataset. Finally, based on the generated haze density and semantic maps, the dual-path aggregation weights are computed to reasonably and effectively aggregate short- and long-range dependencies.

Our contributions are summarized as follows:
\begin{itemize}
\item{We propose DehazeMatic, showing that once effective aggregation is achieved, explicitly and simultaneously modeling short- and long-range dependencies leads to \textit{state-of-the-art} performance on multiple benchmarks.}
\item{We are the first to identify the key factors that determine the relative importance of short- and long-range dependencies in image dehazing. Building on this insight, we propose the CedA module, which dynamically aggregates these dependencies without being restricted to any specific feature extraction operator.}
\item{We further explore the potential of CLIP in haze removal and, for the first time, achieve accurate estimation of haze density maps, without any fine-tuning of the encoder.}
\end{itemize}
 
\section{Related Work}

\subsection{Single Image Dehazing}
Due to spatially variant transmission map and atmospheric light, single image dehazing is a highly ill-posed problem. Early prior-based methods~\cite{he2016deep,fattal2008single,kim2019fast,tan2008visibility,zhu2015fast,berman2018single} achieved dehazing by imposing various assumptions to estimate key parameters in the Atmospheric Scattering Model (ASM)~\cite{narasimhan2002vision}. ASM provides strong theoretical support and interpretability for these methods; however, they often fail when images deviate from statistical laws.

With the rapid advancement of deep learning~\cite{krizhevsky2012imagenet}, various learning-based methods have been proposed, resulting in improved performance. Early methods~\cite{cai2016dehazenet,li2017aod} employ neural networks to estimate key parameters in the ASM and subsequently restore the haze-free images. Later, ASM-independent deep networks~\cite{ren2016single,ren2018gated,liu2019griddehazenet,li2019semi,shao2020domain,dong2020multi,zhang2020pyramid,qin2020ffa,li2020zero,wu2021contrastive,ye2022perceiving,song2023vision,feng2024advancing,yang2024robust,zhang2024depth,chen2024dea,fang2024guided,cong2024semi,wang2024odcr,yang2025unleashing,cui2025eenet} directly estimate clear images or haze residuals. DehazeFormer~\cite{song2023vision} is a representative method that achieves efficient feature extraction through window-based self-attention and several key modifications. However, its inherently limited receptive field constrains its performance potential. To achieve a global receptive field with lower computational overhead, some methods~\cite{shen2023mutual,yu2022frequency} incorporate frequency domain features into image dehazing, offering the additional benefit of more direct learning of high-frequency information. For the same purpose, other researchers~\cite{zheng2024u,zhang2024lmhaze} introduce the Mamba~\cite{gu2023mamba} architecture into dehazing. However, none of these methods address how to better integrate global semantic features with local texture features.

\subsection{CLIP for Low-Level Vision Tasks}
Classic vision-language models like CLIP~\cite{radford2021learning}, aim to learn aligned features in the embedding space from image-text pairs using contrastive learning. Some studies have explored leveraging the rich prior knowledge encapsulated in CLIP to assist with low-level vision tasks.

In All-in-One image restoration, some researchers~\cite{luo2023controlling,ai2024multimodal,jiang2025autodir} use degradation embeddings from the CLIP image encoder to implicitly guide the network in making adaptive responses. In monocular depth estimation, recent studies~\cite{zhang2022can,auty2023learning,hu2024learning} use CLIP to map patches of the input image to specific semantic distance tokens, which are then projected onto a quantified depth bin for estimation. In low-light enhancement, some methods~\cite{liang2023iterative,morawski2024unsupervised} use text-image similarity between the enhanced results and learnable prompt pairs to train the enhancement network.  

Some works also introduce CLIP into image~\cite{wang2024hazeclip} and video~\cite{ren2024triplane} dehazing, but they all use text-image similarity between dehazed results and contrastive prompt sets as regularization to guide the restoration process. In contrast, our method further exploits the potential of CLIP by directly incorporating its latent embeddings into the main network to guide the dehazing process.

\section{Motivational Experiment}
Optimal aggregation of dependencies is essential for a dual-stream network to fully leverage short- and long-range cues for dehazing. To this end, we empirically investigate the key factors that govern their relative importance, thereby enabling more reasonable and effective aggregation.

\subsection{Quantifying the Importance of Dependencies}
Dependency denotes the influence exerted by other tokens on the current token~\cite{bengio1994learning,hochreiter1997long}, and its range can be quantified by the Euclidean distance. For the experiment presented in this section, we train a Transformer model with a global receptive field, where the attention weights assigned by other tokens to the current token in the self-attention mechanism are defined as the importance of the corresponding dependencies.

\subsection{Experimental Design} 
We begin by hypothesizing potential key influencing factors and then verify their validity through quantitative and qualitative results. To this end, we select image tokens with differing characteristics in terms of the hypothesized factors as the current tokens, and examine whether the dependency importance at a fixed distance \textit{varies} accordingly. This setup enables us to reflect the relative importance between short-range and long-range dependencies, as the importance of all other tokens with respect to the current token is normalized using L1 normalization.

\subsection{Experimental Observations} 
\begin{figure}[t]
    \centering
    \includegraphics[width=85mm]{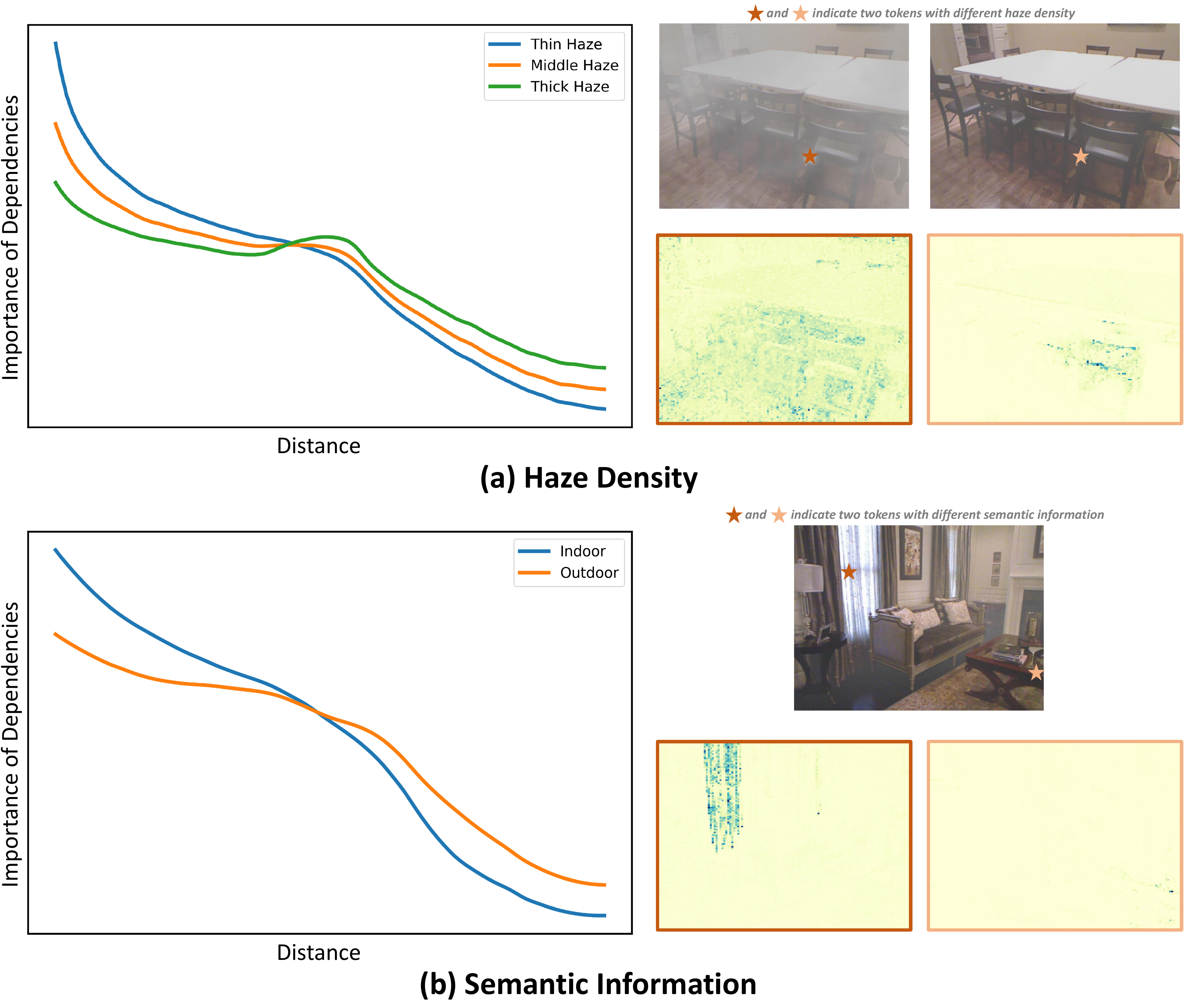}
    \vspace{-16pt}
    \caption{\textbf{Illustration of factors influencing the relative importance between short-range and long-range dependencies}. The left side presents how the quantitative results vary with changes in the characteristics of the hypothesized factors. The horizontal axis represents the Euclidean distance between the current token and all other tokens, \textit{i.e.}, the range of dependency, while the vertical axis indicates the average importance of other dependencies located at the corresponding distance to the current token. The curve is derived from all tokens across all images in the test set. The right side visualizes the importance of other tokens to the chosen tokens (marked with pentagrams). In (a), the tokens are located at the same position and correspond to the same ground truth image but exhibit different haze densities. In (b), the tokens differ in semantic content but are located within the same image under homogeneous haze conditions.}
    \label{fig:demonstrate_different_density_requires_different_receptive}
\end{figure}

\textbf{Haze density} is intuitively regarded as a key contributing factor. To validate this, we synthesize hazy images with varying density levels using the Atmospheric Scattering Model~\cite{narasimhan2002vision} and conduct corresponding experiments. As illustrated in Figure~\ref{fig:demonstrate_different_density_requires_different_receptive}, as the haze becomes denser, the relative importance of long-range dependencies increases, while that of short-range dependencies decreases—and vice versa. Quantitative results further confirm this observation.

\noindent \textbf{Semantic information} is also hypothesized to be an influential factor, as prior work~\cite{huang2020ordnet} indicates that scenes with different levels of complexity require dependencies at varying ranges. To examine this, we conduct experiments on indoor and outdoor images from the RESIDE dataset~\cite{li2018benchmarking}, which generally exhibit distinct semantic characteristics. As shown in Figure~\ref{fig:demonstrate_different_density_requires_different_receptive}, the quantitative results support our hypothesis, while qualitative results further demonstrate that for tokens with different semantic content within the same image, the relative importance of dependencies at different ranges also differs.

Based on these findings, we are the first to propose that in image dehazing, the relative importance of short-range and long-range dependencies is jointly influenced by both \textit{haze density} and \textit{semantic information}.

\section{Method}
\begin{figure*}[ht]
    \centering
    \includegraphics[width=175mm]{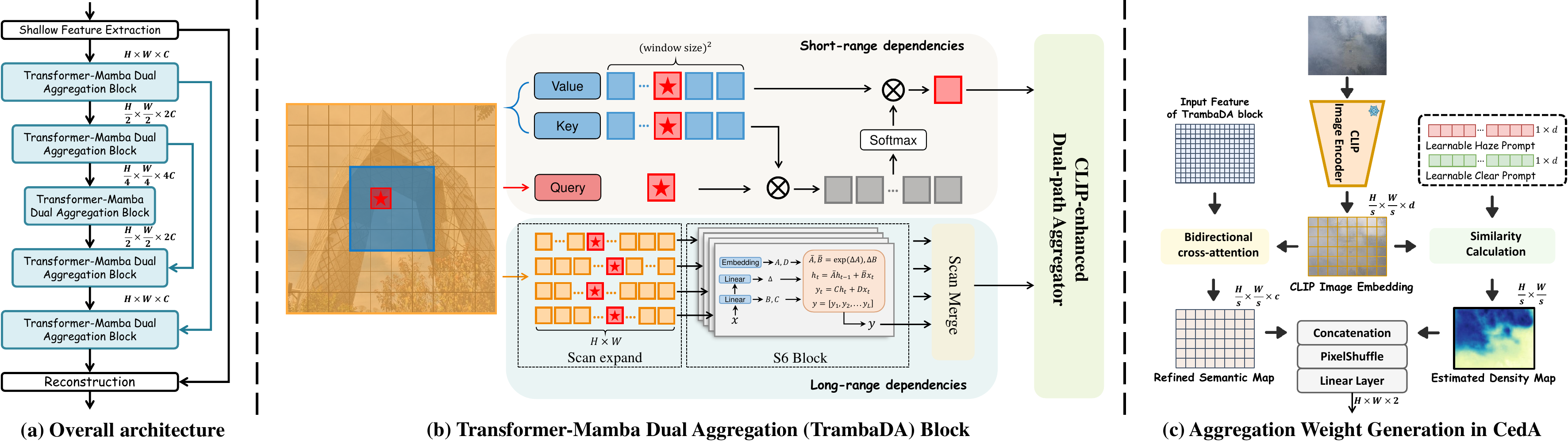}
    \caption{(a) The overall architecture of DehazeMatic. (b) The internal structure of the Transformer-Mamba Dual Aggregation (TrambaDA) block. (c) The process by which the CLIP-enhanced Dual-path Aggregator (CedA) generates the aggregation weights used to fuse the dual-path outputs of the TrambaDA block. To improve computational efficiency, the TrambaDA blocks share the same aggregation weights within each stage.}
    \label{fig:arch}
    \vspace{-10pt}
\end{figure*}

We propose DehazeMatic, as illustrated in Figure~\ref{fig:arch}. It adopts a symmetric encoder-decoder framework in which each stage is constructed using a dual-stream design. To effectively aggregate the outputs from the two streams, we introduce the CLIP-enhanced Dual-path Aggregator, which utilizes the estimated haze density and semantic maps to instruct the aggregation process.

\subsection{Transformer-Mamba Dual Aggregation Block}
The TrambaDA block adopts a dual-stream design to simultaneously capture dependencies at different ranges, aiming to provide complementary restoration information for image tokens with distinct characteristics.

The specific feature extractors for each path are flexible and not the focus of this paper. Therefore, for the sake of design simplicity,  we adopt window-based self-attention ~\cite{liu2021swin} to capture short-range dependencies, with its output denoted as $F_{\mathrm{out\_{short}}}$. To model long-range dependencies, we apply Mamba's S6 block ~\cite{gu2023mamba}, equipped with a four-directional scanning strategy~\cite{liu2024vmamba}, and denote its output as  $F_{\mathrm{out\_{long}}}$. The detailed structure is illustrated in Figure~\ref{fig:arch}(b).

\subsection{CLIP-enhanced Dual-path Aggregator}
Based on the aforementioned insights, CedA is designed to first generate fine-grained haze density and semantic maps, and then produce pixel-level weights accordingly to adaptively aggregate the dual-path outputs, \textit{i.e.},
\begin{equation}
    \label{eq:fusion_operation}
    \begin{split}
    F_{out}=W_{short}F_{\mathrm{out\_{short}}}+W_{long}F_{\mathrm{out\_{long}}}. 
    \end{split}
\end{equation}
$W_{short},W_{long}\in\mathbb{R}^{H \times W}$ are the aggregation weights corresponding to short- and long-range dependencies, respectively, and their generation process is shown in Figure~\ref{fig:arch}(c).

\subsubsection{Estimating Haze Density Map} We begin by defining the estimation pipeline. Inspired by Figure~\ref{fig:clip_can_show_haze_density}, we leverage CLIP to estimate the haze density map. Specifically, given a hazy image $I_\mathrm{haze}\in\mathbb{R}^{H\times W\times 3}$ and a paired prompts describing haze and clear conditions  $T=[T_{\mathrm{haze}},T_{\mathrm{clear}}]$ (\textit{e.g.}, ['hazy image', 'clear image']), we encode them into latent space using a pretrained CLIP model:
\begin{equation}
    \label{eq:clip_encode}
    \begin{split}
    F_{\mathrm{image}}=& \Phi_{\mathrm{image}}(I_\mathrm{haze})\in\mathbb{R}^{H_p \times W_p \times d_c}, \\
    & F_{\mathrm{text}}=\Phi_{\mathrm{text}}(T)\in\mathbb{R}^{2\times d_c},
    \end{split}
\end{equation}
where $\Phi_{\mathrm{text}}$ is the CLIP text encoder, and $\Phi_{\mathrm{image}}$ is the CLIP image encoder \textit{without the final pooling layer}, allowing the generation of a patch-wise density map. $H_p$ and $W_p$ denote the height and width of the encoded image embedding, respectively, while $d_c$ denotes the hidden dimension of CLIP. As shown in Figure~\ref{fig:clip_can_show_haze_density}, as the haze density increases, the similarity scores between $F_{\mathrm{image}}$ and the haze-describing prompt $T_{\mathrm{haze}}$ also increase. Therefore, we treat these similarity scores as the estimated density map $\mathcal{M}_d$:
\begin{equation}
    \label{eq:density_map_generation}
    \begin{split}
    \mathcal{M}_d=\mathrm{Softmax}(\mathrm{sim}(F_{\mathrm{image}},F_{\mathrm{text}}))[:,:,0],
    \end{split}
\end{equation}
where $\mathrm{sim}(\cdot,\cdot)$ is the similarity calculation, and final haze density map $\mathcal{M}_d\in\mathbb{R}^{H_p\times W_p}$.

To optimize estimation performance and reduce reliance on time-consuming prompt engineering, we employ learnable prompt tokens rather than manually predefined prompts to characterize abstract haze and clear conditions.

The training comprises two stages. In Stage 1, we optimize learnable paired prompts using cross-entropy loss, enabling them to preliminarily distinguish between hazy and clear images. Given a hazy and clear image $I_\mathrm{haze},I_\mathrm{clear}\in\mathbb{R}^{H\times W\times 3}$, the first-stage loss $\mathcal{L}_1$ is defined as: 
\begin{equation}
    \label{eq:learn_prompt_pair_l1}
    \begin{split}
    \mathcal{L}_1=-(y*\log(\hat{y})+(1-y)*\log(1-\hat{y})), \\
    \hat{y}=\frac{e^{cos(\mathrm{Pool}(\Phi_{\mathrm{image}}(I)),\Phi_{\mathrm{text}}(T_\mathrm{clear}))}}{\sum_{i\in\{\mathrm{haze},\mathrm{clear}\}}e^{cos(\mathrm{Pool}(\Phi_{\mathrm{image}}(I)),\Phi_{\mathrm{text}}(T_i))}},
    \end{split}
\end{equation}
where $I\in\{I_\mathrm{haze},I_\mathrm{clear}\}$ and $y$ is the label of the current image, 0 is for hazy image $I_\mathrm{haze}$ and 1 is for clear image $I_\mathrm{clear}$. Since $\Phi_{\mathrm{image}}$ is the CLIP image encoder without the final pooling layer, we apply a global pooling layer $\mathrm{Pool}()$ to obtain the latent image embedding with a shape of $\mathbb{R}^{1\times d_c}$.

In Stage 2, the objective is to predict haze density more accurately. The most straightforward and effective optimization approach is regression. However, existing datasets lack ground-truth density maps corresponding to hazy images. To address this, we construct triplet data $\{I_\mathrm{haze},I_\mathrm{clear},I_\mathrm{density}\}$ based on Atmospheric Scattering Model~\cite{narasimhan2002vision}. The second-stage loss $\mathcal{L}_2$ is then defined as:
\begin{equation}
    \label{eq:learn_prompt_pair_l2}
    \mathcal{L}_2 =
    \begin{cases}
        \alpha_1 \mathrm{MSE}(\mathcal{M}_{d},I_\mathrm{density}) + \alpha_2 \mathcal{L}_1, & y=0 \\
        \mathcal{L}_1, & y=1
    \end{cases}
\end{equation}
where $\mathrm{MSE}()$ is the mean absolute error, and $\mathcal{M}_{d}$ is obtained from $I_\mathrm{haze}$ and the learnable paired prompts $T=[T_{\mathrm{haze}},T_{\mathrm{clear}}]$ according to Equation~\ref{eq:density_map_generation}. $\alpha_1$ and $\alpha_2$ are the weights of different loss functions. Training is lightweight since there is no need to fine-tune the CLIP image encoder.

\begin{figure}[t]
    \centering
    \includegraphics[width=85mm]{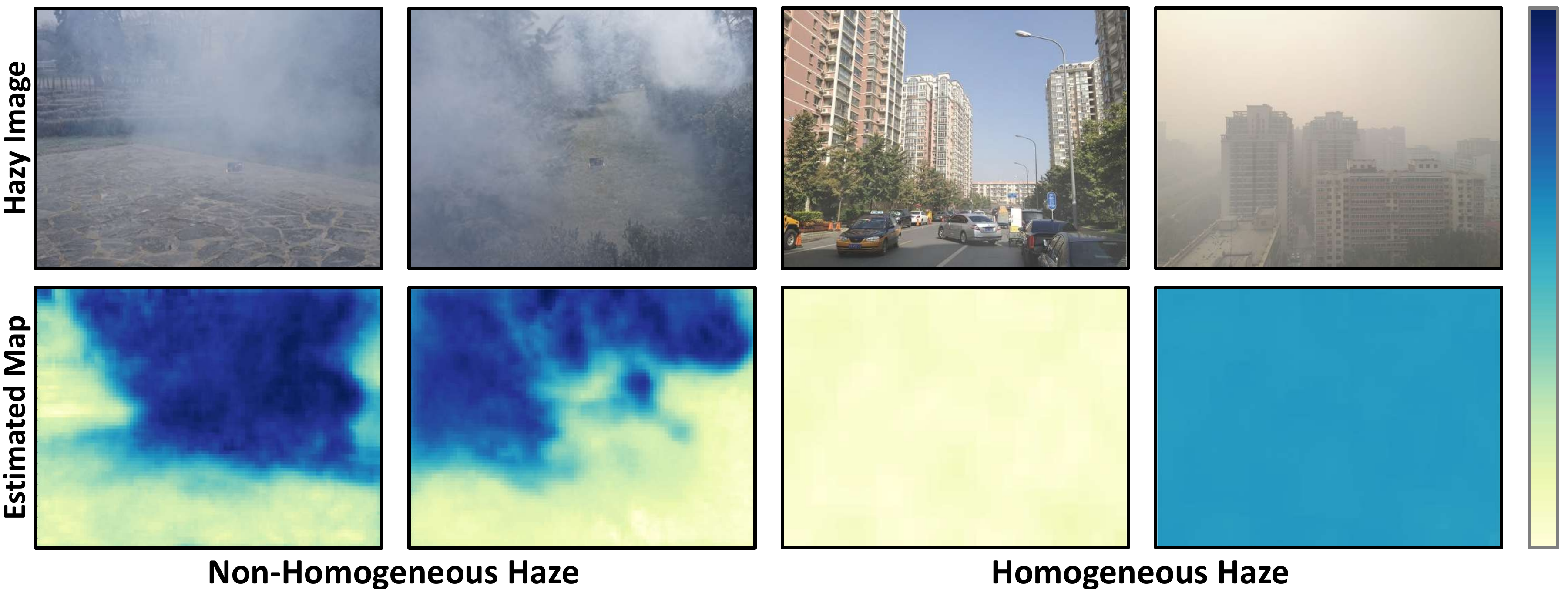}
    \vspace{-18pt}
    \caption{\textbf{Illustration of estimated haze density maps}. The proposed method can identify the relative haze density not only in different regions of non-homogeneous haze images but also between two homogeneous haze images.}
    \label{fig:estimated_density_map}
\end{figure}

Finally, we use the learned paired prompts $[T_{\mathrm{haze}},T_{\mathrm{clear}}]$ to generate the estimated patch-wise haze density map $\mathcal{M}_d$. Some examples are shown in Figure~\ref{fig:estimated_density_map}, demonstrating that our method is an effective and universal approach applicable to both homogeneous and non-homogeneous haze.

\subsubsection{Obtaining Refined Semantic Map}
Considering that the pretrained image encoder of CLIP is obtained through a classification pretext task, each location in the feature map before pooling captures regional semantic information~\cite{zhang2022can}. Therefore, we reuse $F_{\mathrm{image}}$ from Equation~\ref{eq:clip_encode} and regard it as semantic information. This approach not only provides high-level information but also avoids introducing additional computational overhead.

However, using only $F_{\mathrm{image}}$ gives suboptimal results, primarily because it is an abstract representation and lacks low-level semantic details. To fix this, we incorporate the input features of the TrambaDA blocks, $F_{\mathrm{input}}$, to supplement $F_{\mathrm{image}}$ and introduce a bidirectional cross-attention mechanism to align their semantic information across different scales, thereby yielding a refined semantic map $\mathcal{M}_s$:
\begingroup
\begin{equation}
    \label{eq:cross_attention_input_to_image}
    \begin{split}
    \mathcal{M}_s=\mathrm{Linear}&(\text{Attention}(Q_{image},K_{input},V_{input}), \\
    &(\text{Attention}(Q_{input},K_{image},V_{image})).
    \end{split}
\end{equation}
\endgroup
$Q_i$, $K_i$ and $V_i$ represent the query, key, and value derived from $F_\mathrm{i}$ after channel reduction or adaptive pooling, where $\mathrm{i}\in\{\mathrm{image},\mathrm{input}\}$. $\text{Attention}(\cdot)$ is self-attention operation.

\subsubsection{Generating Dual-path Aggregation Weights}
Finally, the estimated haze density map $\mathcal{M}_d$ and the obtained refined semantic map $\mathcal{M}_s$ are jointly used to generate the dual-path aggregation weights, as follows:
\begin{equation}
    \label{eq:weights_generation}
    \begin{split}
    W_{short},W_{long} & =\mathrm{Split}(\mathrm{Linear}(\mathrm{Interpolate}( \\
    & \quad \quad \quad \quad \mathrm{Concat}(\mathcal{M}_d,\mathcal{M}_s)))).
    \end{split}
\end{equation}
Here, $\mathrm{Concat}(\cdot)$ is concatenation operation,  $\mathrm{Interpolate}(\cdot)$ represents an interpolation operation used to resize $\mathcal{M}_d$ and $\mathcal{M}_s$ to match the dimensions of $F_{\mathrm{out\_{short}}}$ and $F_{\mathrm{out\_{long}}}$, $\mathrm{Linear}(\cdot)$ refers to a projection layer, and $\mathrm{Split}(\cdot)$ indicates a channel-wise splitting operation. The aggregation is subsequently performed using Equation~\ref{eq:fusion_operation}.
\section{Experiments}
\subsection{Experiment Setup}
\noindent \textbf{Training Details.} DehazeMatic is implemented with PyTorch on NVIDIA A100 GPUs. We use Adam~\cite{kingma2014adam} optimizer with default parameters ($\beta_1=0.9$, $\beta_2=0.99$) and a cosine annealing strategy~\cite{loshchilov2016sgdr} with restarts. The initial learning rate is set to $2\times10^{-4}$, gradually decreasing to $2 \times10^{-6}$. We train the homogeneous haze dataset for 200 epochs and the non-homogeneous haze dataset for 400 epochs. The images are randomly cropped to a size of 256$\times$256 and augmented with flipping. We use L1  loss and perceptual loss~\cite{johnson2016perceptual} to supervise dehazing process.

\noindent \textbf{Datasets.} We evaluate DehazeMatic on both synthetic and real-world datasets. The synthetic datasets encompass both homogeneous and non-homogeneous haze scenarios. For homogeneous haze, we utilize RESIDE dataset~\cite{li2018benchmarking}, which includes two training subsets: Indoor Training Set (ITS), containing 13,990 paired indoor images, and Outdoor Training Set (OTS), containing 313,950 paired outdoor images. Evaluation is conducted on the corresponding subsets of the Synthetic Objective Testing Set (SOTS). For non-homogeneous haze, we use the NH-HAZE~\cite{ancuti2020nh} and Dense-Haze~\cite{ancuti2019dense} datasets, both generated using a professional haze generator to simulate real-world hazy conditions. Each of these datasets comprises 55 paired images, of which the last 5 are used for testing and the remaining 50 for training. For real-world evaluation, we use the RTTS dataset~\cite{li2018benchmarking}, which contains 4,322 real-world hazy images.

\subsection{Comparison with SOTA Methods}
\begin{table*}[t]
    \centering
    \setlength{\abovecaptionskip}{0cm}
    \caption{\textbf{Quantitative comparisons of various methods on multiple synthetic dehazing benchmarks}. The best results and the second best results are in \textbf{bold} and \underline{underline}, respectively.}
    \label{tab:quantitative_results}
    \begin{center}
    \renewcommand\arraystretch{}
    {
        \resizebox{0.95\textwidth}{!}{
        \begin{tabular}{l|cccccc|cccccc|rr}
        \multirow{3}{*}{Methods} & \multicolumn{6}{c|}{Homogeneous Haze} & \multicolumn{6}{c|}{Non-homogeneous Haze} & \multicolumn{2}{c}{\multirow{2}{*}{Overhead}}\\
        & \multicolumn{2}{c}{SOTS-Outdoor} & \multicolumn{2}{c}{SOTS-Indoor} & \multicolumn{2}{c|}{Average} & \multicolumn{2}{c}{NH-Haze} & \multicolumn{2}{c}{Dense-Haze} & \multicolumn{2}{c|}{Average} & &  \\
        & PSNR$\bm{\uparrow}$ & SSIM$\bm{\uparrow}$ & PSNR$\bm{\uparrow}$ & SSIM$\bm{\uparrow}$ & PSNR$\bm{\uparrow}$ & SSIM$\bm{\uparrow}$ & PSNR$\bm{\uparrow}$ & SSIM$\bm{\uparrow}$ & PSNR$\bm{\uparrow}$ & SSIM$\bm{\uparrow}$ & PSNR$\bm{\uparrow}$ & SSIM$\bm{\uparrow}$ & \#Param & MACs \\
        \toprule
        
        DCP~\cite{he2010single} & 19.14 & 0.861 & 16.61 & 0.855 & 17.88 & 0.858 & 10.57 & 0.520 & 10.06 & 0.385 & 10.32 & 0.453 & - & - \\
        AOD-Net~\cite{li2017aod} & 24.14 & 0.920 & 20.51 & 0.816 & 22.33 & 0.868 & 15.40 & 0.569 & 13.14 & 0.414 & 14.27 & 0.492 & 1.76K & 0.12G \\
        GridDehazeNet~\cite{liu2019griddehazenet} & 30.86 & 0.982 & 32.16 & 0.984 & 31.51 & 0.983 & 18.33 & 0.667 & 14.96 & 0.533 & 16.65 & 0.600 & 0.96M & 21.55G \\
        FFA-Net~\cite{qin2020ffa} & 33.57 & 0.984 & 36.39 & 0.989 & 34.98 & 0.987 & 19.87 & 0.692 &  16.09 &  0.503 & 17.98 & 0.598 & 4.46M & 288.86G \\
        DeHamer~\cite{guo2022image} & 35.18 & 0.986 & 36.63 & 0.988 & 35.91 & 0.987 & 20.66 & 0.684 & 16.62 & 0.560 & 18.64 & 0.622 & 132.45M & 59.25G \\
        SRDefog~\cite{jin2022structure} & - & - & - & - & - & - & 20.99 & 0.610 & 16.67 & 0.500 & 18.83 & 0.555 & 12.56M & 24.18M \\
        MAXIM-2S~\cite{tu2022maxim} & 34.19 & 0.985 & 38.11 & 0.991 & 36.15 & 0.988 & - & - & - & - & - & - & 14.10M & 216.00G \\
        SGID-PFF~\cite{bai2022self} & 30.20 & 0.975 & 38.52 & 0.991 & 34.36 & 0.983 & - & - & - & - & - & - & 13.87M & 156.67G \\
        PMNet~\cite{ye2022perceiving} & 34.74 & 0.985 & 38.41 & 0.990 & 36.58 & 0.988 & 20.42 & 0.730 & 16.79 & 0.510 & 18.61 & 0.620 & 18.90M & 81.13G \\
        MB-TaylorFormer-B~\cite{qiu2023mb} & \underline{37.42} & 0.989 & 40.71 & 0.992 & 39.07 & 0.991 & - & - & 16.66 & 0.560 & - & - & 2.68M & 38.50G \\
        MITNet~\cite{shen2023mutual} & 35.18 & 0.988 & 40.23 & 0.992 & 37.71 & 0.990 & \underline{21.26} & 0.712 & \underline{16.97} & 0.606 & \underline{19.12} & 0.659 & 2.73M & 16.42G \\
        DehazeFormer~\cite{song2023vision} & 34.29 & 0.983 & 38.46 & 0.994 & 36.38 & 0.989 & 20.31 & 0.761 & 16.66 & 0.595 & 18.49 & 0.595 & 4.63M & 48.64G \\
        SCANet~\cite{guo2023scanet} & - & - & - & - & - & - & 19.52 & 0.649 & 15.35 & 0.508 & 17.44 & 0.579 & 2.39M & 258.63G \\
        DEANet~\cite{chen2024dea} & 36.03 & 0.989 & 40.20 & 0.993 & 38.12 & 0.991 & 20.84 & \underline{0.801} & 16.73 & 0.602 & 18.79 & 0.702 & 3.65M & 32.23G \\
        UVM-Net~\cite{zheng2024u} & 34.92 & 0.984 & 40.17 & \textbf{0.996} & 37.55 & 0.990 & - & - & - & - & - & - & 19.25M &  173.55G \\
        OKNet~\cite{cui2024omni} & 35.45 & 0.992 & 37.59 & 0.994 & 36.52 & 0.993 & 20.29 & 0.800 & 16.85 & \underline{0.620} & 18.57 & \underline{0.710} & 4.42M & 39.54G \\
        DCMPNet~\cite{zhang2024depth} & 36.56 & \underline{0.993} & \textbf{42.18} & \textbf{0.996} & \underline{39.37} & 0.995 & - & - & - & - & - & - & 18.59M & 80.42G\\
        \midrule
        \rowcolor{gray!15} DehazeMatic & \textbf{38.21} & \textbf{0.995}  & \underline{41.50} & \textbf{0.996} & \textbf{39.86} & \textbf{0.996} & \textbf{21.47} & \textbf{0.806}  & \textbf{17.28} & \textbf{0.629}  & \textbf{19.38} & \textbf{0.718} & 4.58M & 35.50G \\
        \bottomrule
        \end{tabular}}
    }
  \end{center}
\end{table*}

\begin{table}[t]
    \begin{center}
    \setlength{\abovecaptionskip}{0cm}
    \caption{\textbf{Quantitative comparison on the real-world dehazing dataset RTTS.} The best results are in \textbf{bold}.}
    \label{tab:RTTS}
    \resizebox{\linewidth}{!}{
    {
        \begin{tabular}{l|ccc}
        \toprule
        Methods & FADE~$\boldsymbol{\downarrow}$ & BRISQUE~$\boldsymbol{\downarrow}$ & NIMA~$\boldsymbol{\uparrow}$ \\
        \midrule
        PSD~\cite{chen2021psd} & 0.920 & 27.713 & 4.598 \\
        D4~\cite{yang2024robust} & 1.358 & 33.210 & 4.484 \\
        DGUN~\cite{mou2022deep} & 1.111 & 27.968 & 4.653 \\
        RIDCP~\cite{wu2023ridcp} & 0.944 & 17.293 & 4.965 \\
        CORUN~\cite{fang2024real} & 0.824 & 11.956 & 5.342 \\
        SGDN~\cite{fang2025guided} & 0.873 & 11.549 & 5.128 \\
        \midrule
        \rowcolor{gray!15} DehazeMatic & \textbf{0.796} & \textbf{11.435} & \textbf{5.510} \\
        \bottomrule
        \end{tabular}
    }}
    \end{center}
\end{table}

\begin{figure*}[ht]
    \centering
    \includegraphics[width=175mm]{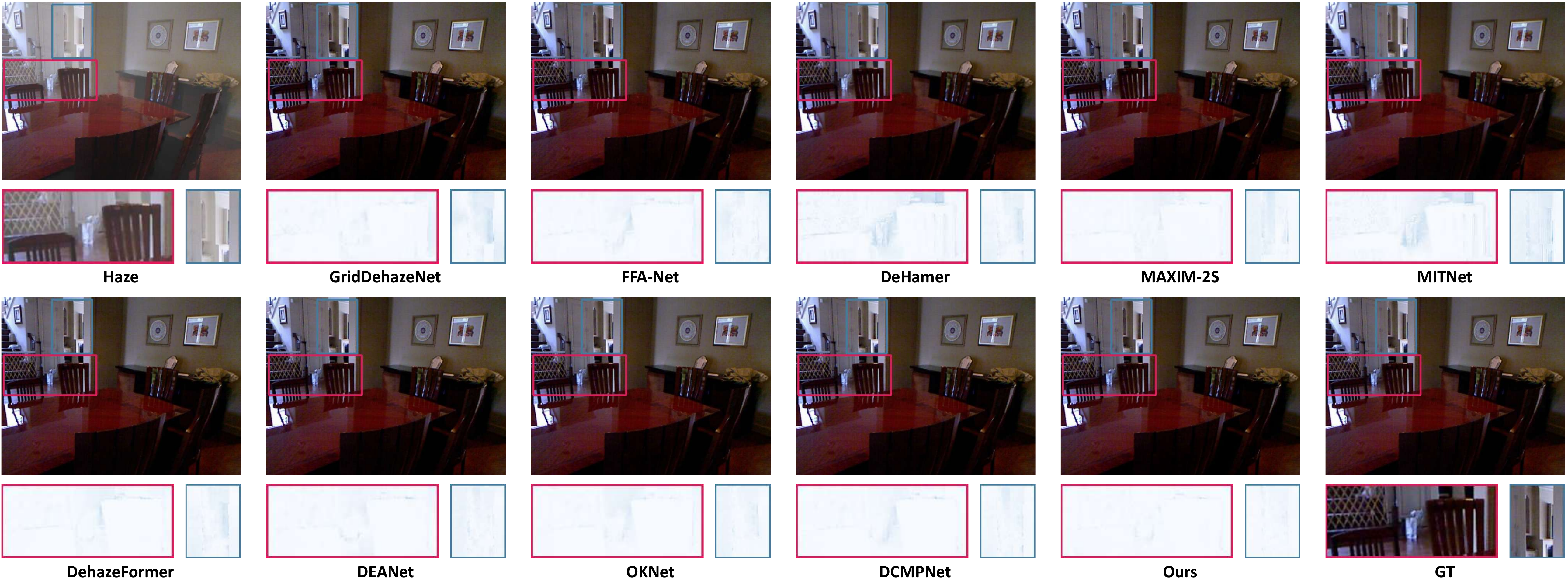}
    \caption{\textbf{Visual comparisons on homogeneous hazy images}. The bottom displays zoomed-in error maps of selected regions, where darker blue indicates larger restoration errors. Please zoom in to view.}
    \label{fig:vis_comparison_RESIDE_IN}
\end{figure*}

\begin{figure*}[ht]
    \centering
    \includegraphics[width=175mm]{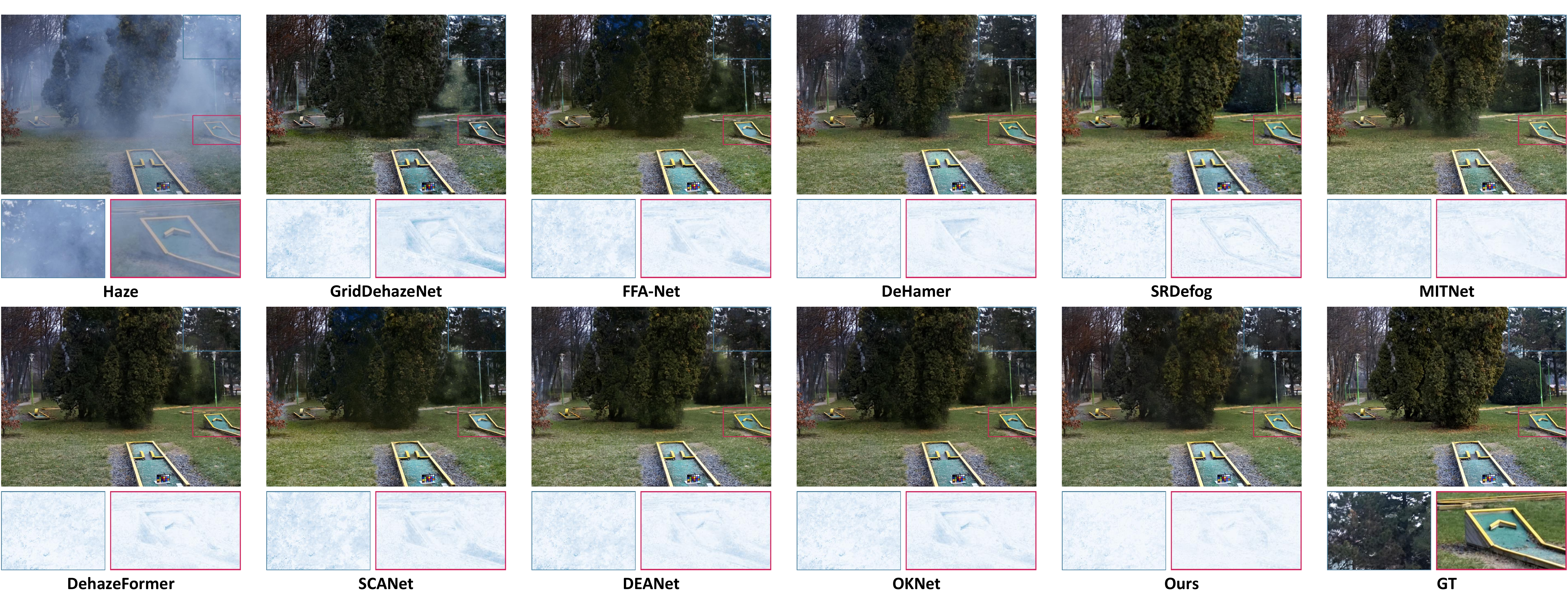}
    \caption{\textbf{Visual comparisons on non-homogeneous hazy images}. The bottom displays zoomed-in error maps of selected regions, where darker blue indicates larger restoration errors. Please zoom in to view.}
    \label{fig:visual_result_NHHaze}
\end{figure*}

\begin{figure}[t]
    \centering
    \includegraphics[width=85mm]{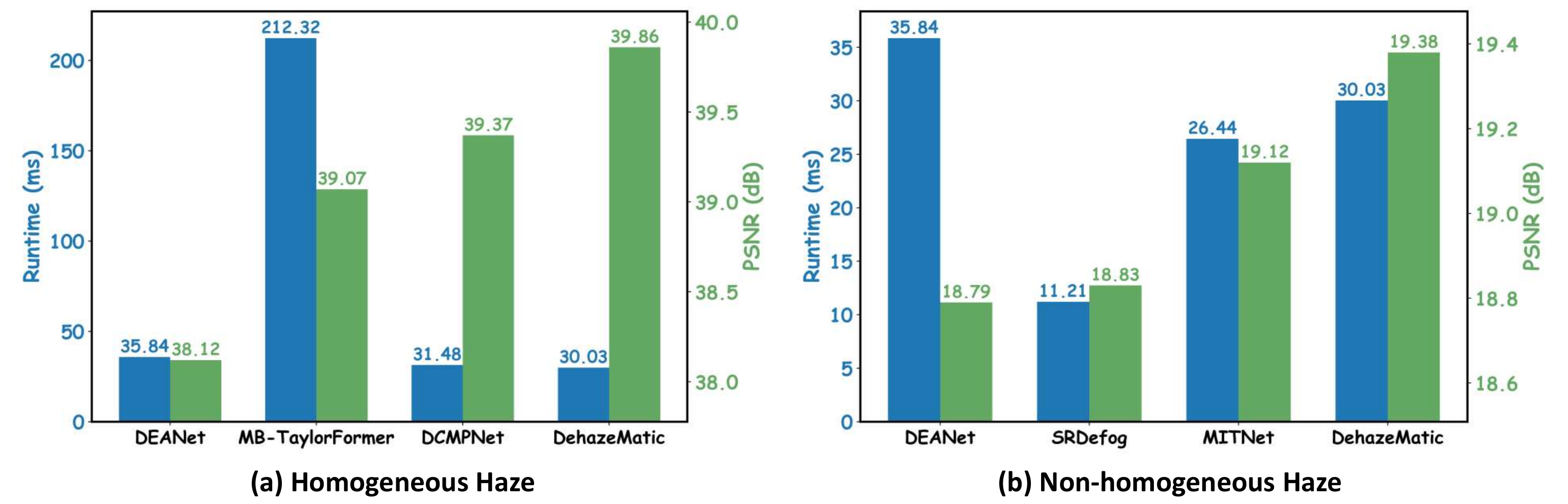}
    \caption{\textbf{Illustration of the trade-off between performance and runtime}. The blue bar denotes runtime, while the green bar indicates performance. A larger gap, where the green bar exceeds the blue bar, signifies a better trade-off.}
    \label{fig:trade-off}
\end{figure}

\subsubsection{Performance on Synthetic Homogeneous Haze} As shown in Table~\ref{tab:quantitative_results}, both DCMPNet~\cite{zhang2024depth} and MB-TaylorFormer~\cite{qiu2023mb} are competitive approaches; however, our DehazeMatic achieves the best overall performance. DCMPNet utilizes depth information as guidance, while MB-TaylorFormer employs linear self-attention for feature extraction. These results demonstrate the superiority of jointly leveraging haze density maps and semantic maps for guidance, as well as simultaneously capturing both short- and long-range dependencies. A visual comparison is shown in Figure~\ref{fig:vis_comparison_RESIDE_IN}. Notably, only our method yields minimal restoration errors at object boundaries.

\subsubsection{Performance on Synthetic Non-homogeneous Haze} As shown in Table~\ref{tab:quantitative_results}, while MITNet~\cite{shen2023mutual} and OKNet~\cite{cui2024omni} achieve competitive performance in PSNR and SSIM, respectively, our DehazeMatic consistently outperforms them by achieving SOTA results across all metrics. The error maps in Figure~\ref{fig:visual_result_NHHaze} demonstrate that our method preserves the most intricate details in the blue-boxed tree region and yields the smallest errors along the object boundaries in the red-boxed target area.

\subsubsection{Performance on Real-world Haze} To evaluate the practicality and generalization ability of our model in real-world scenarios, we conduct experiments on the real-world haze dataset RTTS~\cite{li2018benchmarking}, comparing it with SOTA methods specifically designed for real-world dehazing. As shown in Table~\ref{tab:RTTS}, our method outperforms all others in terms of no-reference metrics, demonstrating its superiority.

\subsubsection{Trade-off between Performance and Runtime} To evaluate the efficiency of our model, we measure its runtime on an NVIDIA A100 GPU and compare the performance and runtime trade-off with the three most competitive methods, as shown in Figure~\ref{fig:trade-off}. Although our method incorporates CLIP~\cite{radford2021learning}, the additional inference time it introduces is negligible—only 3.8 ms. As a result, the overall runtime of our model is merely 30.03 ms, achieving the best trade-off. Moreover, our method reaches 33 frames per second (FPS), satisfying real-time processing requirements.

\subsection{Ablation Studies}
We conduct extensive ablation studies to demonstrate the effectiveness of each component. For a fair comparison, we tune the hyperparameters of variant models to ensure that their overhead matches that of DehazeMatic.

\begin{table}[t]
    \begin{center}
    \setlength{\abovecaptionskip}{0cm}
    \caption{Ablation study of the explicit coexistence of short- and long-range dependencies.}
    \label{tab:ablation_study_dual_path_design}
    \resizebox{\linewidth}{!}{
    {
        \begin{tabular}{l|cc|cc}
        \toprule
        \multirow{2}{*}{Setting} & \multicolumn{2}{c|}{SOTS-Indoor} & \multicolumn{2}{c}{NH-Haze} \\
        & PSNR$\bm{\uparrow}$ & SSIM$\bm{\uparrow}$ & PSNR$\bm{\uparrow}$ & SSIM$\bm{\uparrow}$ \\
        \midrule
        Only Short Dependencies & 38.41 & 0.992 & 20.40 & 0.771 \\ 
        Only Long Dependencies & 39.78 & 0.994 & 20.56 & 0.779 \\
        \rowcolor{gray!15} DehazeMatic & \textbf{41.50} & \textbf{0.996} & \textbf{21.47} & \textbf{0.806} \\
        \midrule
        Only Convolution & 38.29 & 0.992 & 20.33 & 0.767 \\ 
        Only Linear Attention & 40.06 & 0.994 & 20.65 & 0.780 \\ 
        Conv. \& Linear Attention & \textbf{41.41} & \textbf{0.996} & \textbf{21.36} & \textbf{0.803} \\ 
        \bottomrule
        \end{tabular}
    }}
    \end{center}
\end{table}
\subsubsection{Explicit Coexistence of Short- and Long-range Dependencies} To validate the effectiveness of the dual-stream design, we construct variant models by removing either the path responsible for capturing long-range dependencies or the one for short-range dependencies. The corresponding results are presented in the first three rows of Table~\ref{tab:ablation_study_dual_path_design}.

Furthermore, to validate the generality of our dual-stream design and show that it does not depend on specific feature extractors in each path, we construct a variant model that employs convolution~\cite{zheng2023curricular} (\textit{i.e.}, short-range) and linear self-attention~\cite{qiu2023mb} (\textit{i.e.}, long-range), and repeat the above experiments. The corresponding quantitative results are reported in the last three rows of Table~\ref{tab:ablation_study_dual_path_design}. These results confirm that the integration of complementary short- and long-range dependencies through the dual-stream design substantially improves image dehazing performance.

\begin{table}[t]
    \begin{center}
    \setlength{\abovecaptionskip}{0cm}
    \caption{Ablation study of the CedA.}
    \label{tab:ablation_study_ceda}
    \resizebox{\linewidth}{!}{
    {
        \begin{tabular}{lr|cc|cc}
        \toprule
        \multicolumn{2}{c|}{Setting} & \multicolumn{2}{c|}{SOTS-Indoor} & \multicolumn{2}{c}{NH-Haze} \\
        & & PSNR$\bm{\uparrow}$ & SSIM$\bm{\uparrow}$ & PSNR$\bm{\uparrow}$ & SSIM$\bm{\uparrow}$ \\
        \midrule
        \multirow{2}{*}{(a) Remove CedA} & Addition & 39.55 & 0.993 & 20.71 & 0.781 \\
        & Concatenation & 39.80 & 0.994 & 20.74 & 0.784 \\
        \midrule
        \multirow{2}{*}{(b) Overall Design} & W/o Density Map & 40.88 & 0.995 &  20.91 & 0.792 \\
        & W/o Semantic Map  & 41.02 & 0.995 & 21.23 & 0.794 \\
        \midrule
        & Transmission Map  & 40.97 & 0.995 & 21.27 & 0.797  \\
        (c) Density Map & Depth Map  & 41.16 & 0.996 & 21.28 & 0.795 \\
        & Predefined Prompts & 40.24 & 0.994 & 20.82 & 0.789 \\
        \midrule
        \multirow{2}{*}{(d) Sematic Map} & W/o High-level & 41.12 & 0.995 & 21.30 & 0.800 \\
        & W/o Low-level & 41.19 & 0.996 & 21.34 & 0.802\\
        \midrule
        \rowcolor{gray!15} \multicolumn{2}{c|}{DehazeMatic} & \textbf{41.50} & \textbf{0.996} & \textbf{21.47} & \textbf{0.806} \\
        \bottomrule
        \end{tabular}
    }}
    \end{center}
\end{table}
\subsubsection{CLIP-enhanced Dual-path Aggregator (CedA)}
As shown in Table~\ref{tab:ablation_study_ceda}, we conduct extensive ablation studies on the CedA to verify the effectiveness of each component.

\textit{(a)} We replace the CedA with addition and concatenation operations to verify the necessity of adaptively aggregating short- and long-range dependencies in the dual paths.

\textit{(b)} We then verify the importance of jointly using haze density and semantic information map for aggregation guidance by removing each component individually.

\begin{figure}[t]
    \centering
    \includegraphics[width=82mm]{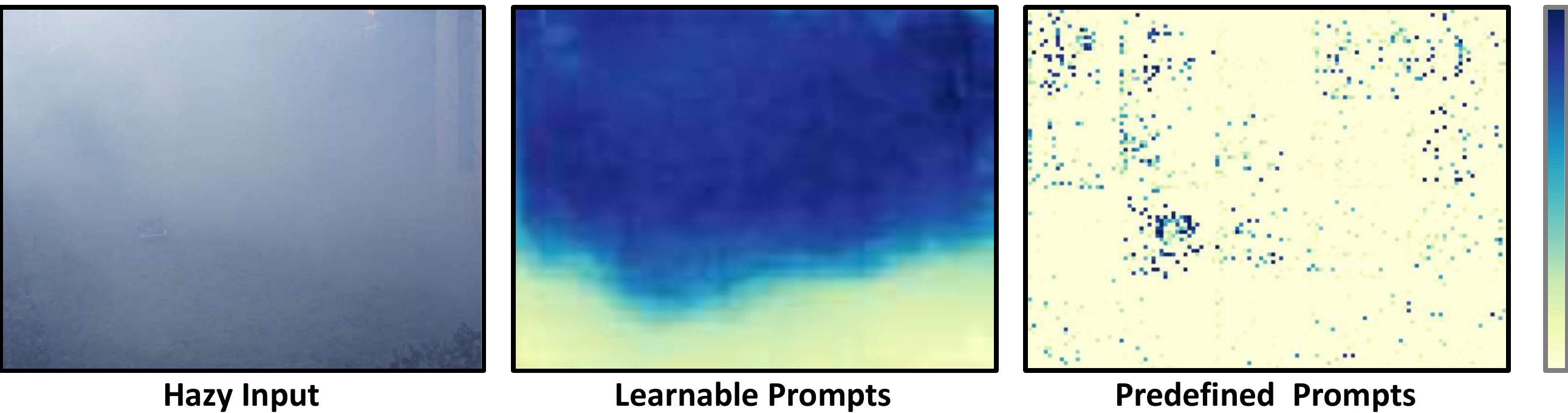}
    \caption{Visual comparisons of haze density maps estimated by learned prompts and manually predefined prompts.}
    \label{fig:comparison_of_learnable_prompts_and_manul_prompts}
\end{figure}
\textit{(c)} Next, we investigate the estimated haze density map. First, we verify the rationale for using the density map rather than other commonly used guidance signals in dehazing (\textit{i.e.}, the transmission map from DCP~\cite{he2010single} and the depth map from Depth Anything~\cite{yang2024depth}) to guide the aggregation. The results indicate that haze density is a more suitable influencing factor.

Furthermore, we use manually predefined prompts to estimate the haze density map to verify the effectiveness of using learnable prompts. As shown in Table~\ref{tab:ablation_study_ceda}, predefined prompts lead to a significant performance drop, even falling behind the approach that relies solely on semantic maps for aggregation guidance. In addition, as illustrated in Figure~\ref{fig:comparison_of_learnable_prompts_and_manul_prompts}, predefined prompts are entirely incapable of generating valid patch-wise haze density map.

\textit{(d)} Finally, we validate the necessity of each type of information in the refined semantic map by ablating either high-level semantic information (\textit{i.e.}, from CLIP) or low-level semantic information (\textit{i.e.}, input of Tramba blocks).
\vspace{-2pt}
\section{Conclusion}
We propose DehazeMatic, demonstrating that explicitly capturing complementary short- and long-range dependencies, along with effective aggregation, can achieve SOTA performance and holds potential to inspire future research. To ensure that dependencies across different ranges contribute optimally, we first identify two key factors influencing their relative importance: haze density and semantic information. Guided by this insight, we introduce the CLIP-enhanced Dual-path Aggregator, which generates both maps to instruct the aggregation process. To the best of our knowledge, we are the first to accurately estimate fine-grained haze density maps, further unlocking CLIP’s potential in dehazing.

\bibliography{aaai2026}

\end{document}